\documentclass{article}

     \usepackage[final]{neurips_2018}

\usepackage[utf8]{inputenc} 
\usepackage[T1]{fontenc}    
\usepackage{hyperref}       
\usepackage{url}            
\usepackage{booktabs}       
\usepackage{amsfonts}       
\usepackage{nicefrac}       
\usepackage{microtype}      
\usepackage{graphicx}
\usepackage{bold-extra}

\title{Challenges and Thrills of Legal Arguments}

\author{%
  Anurag Pallaprolu\\
  UC Santa Barbara\\
  Santa Barbara, CA, 93106 \\
  \texttt{apallaprolu@ucsb.edu} \\
  \And
   Radha Vaidya \\
  UC Santa Barbara\\
  Santa Barbara, CA, 93106 \\
  \texttt{radhavaidya@ucsb.edu} \\
   \And
   Aditya Swaroop Attawar \\
  UC Santa Barbara\\
  Santa Barbara, CA, 93106 \\
  \texttt{adityaswaroop@ucsb.edu} \\
}

\begin{document}

\maketitle

\begin{abstract}
State-of-the-art attention based models, mostly centered around the transformer architecture, solve the problem of sequence-to-sequence translation using the so-called scaled dot-product attention. While this technique is highly effective for estimating inter-token attention, it does not answer the question of inter-sequence attention when we deal with conversation-like scenarios. We propose an extension, HumBERT, that attempts to perform continuous contextual argument generation using locally trained transformers.
\end{abstract}

\section{Prelude}
Legal arguments extensively involve text and speech. For any statement to be generated in the legal context, we require a comprehensive knowledge of prior information regarding the case, the arguments that have previously been made as well as external knowledge and facts. This makes the industry heavily data-driven, and consequently, an interesting domain for the application of machine learning models to automate and improve contextual coherence of arguments.
\vspace{3 mm}
\\
The application of machine learning in the legal industry could imply categorizing certain types of documents or generating plausible arguments to help lawyers with their cases. With efficient algorithms, lawyers can avoid repetitive work and instead focus on complex, higher-value analysis to solve their clients’ legal problems, resulting in a substantial saving of time and effort. This redefines the scope of what lawyers and firms can achieve, allowing them to take on cases which would have been too time-consuming or too expensive for the client if they were to be conducted manually.
\vspace{3 mm}
\\
Since law is essentially expressed in vernacular, Natural Language Processing (NLP) is a crucial component in understanding and prediction of information and contexts. NLP aids in solving computational problems like information retrieval, information extraction, speech recognition and question-answering. One of the most efficient techniques to achieve accurate prediction of a sequence of words is Neural Machine Translation, which has subsequently led to the development of the domain of Transfer Learning.

\section{Transfer Learning}
There are several drawbacks when using algorithms that perform supervised learning (SL), unsupervised learning (UL) or reinforcement learning (RL), especially when it comes to robustness to new/unknown inputs. Supervised learning is not generalizable as it breaks down when we do not have sufficient labeled data to train a reliable model. Unsupervised learning is generalizable only when a stable prior distribution exists and it too shows fragility when we are presented with an outlier. Although reinforcement learning is generalizable, it is computationally intensive since it addresses the task of selecting actions to maximize the reward function through state observation and interaction with the environment.
\vspace{3 mm}
\\
Transfer Learning [11] is an attempt to use one supervised learning model to work on another related setting with minimal re-training. It allows us to deal with scenarios where a model can be trained using a similar "pre-trained" one by leveraging the existing labeled data of some related task or domain. This gained knowledge is stored for solving the source task in the source domain and we apply it to our problem of interest. For example: in our case we are dealing with language modeling on a legal corpus which is also equivalently a valid English text dataset. Hence, two models that show breadth-level similarity could have a common parent domain and due to this advantage, transfer learning is witnessing a sharp rise in its usage across different applications.
\vspace{3 mm}
\\
Transfer Learning is a key aspect of this project since the generation of legal arguments involves learning from prior actions to make more informed and coherent statements. In the NLP domain, transfer learning entails making use of correlations between words generated previously with the context currently under discussion. This has enabled prediction algorithms to become more accurate and efficient.

\section{Problem Statement}
It has always been challenging to generate coherent long-form text. Even with the breakthroughs in Neural Machine Translation algorithms in achieving local dependencies within sentences, we still have a long way to go before we can fully capture global dependencies within a conversation. The concept of 'attention' introduced in [21, 22] brought a fresh perspective to the field of language modeling. This concept has been extensively studied in intra-sentence framework but in recent times there have been a few instances of focused research directed towards conversational coherence [5, 6].
\vspace{3mm}
\\
Our main contribution in this paper is to build a dialog agent by proposing a unique architecture that ensures intra-sentence as well as inter-sentence coherence and cohesion. Such a model can prove useful in many industry-based as well as everyday tasks. In this paper, we propose a use-case for such a model in the legal industry, which is highly text-data intensive.

\section{Preliminaries}
It is imperative to understand the historical evolution of similar ideas in the sphere of natural language processing to make sense of our extension. We will briskly step through some of the recent advancements before taking a look at the attention revolution ushered in by the transformer model.
\subsection{Early Work}
Natural Language Understanding (NLU) had its origins in techniques that did not use any explicit hypothesis classes in their respective SL settings. Latent Semantic Analysis (LSA) [1, 2], for instance, operates on extracting dominant tokens from a corpus of documents by working with the Term Frequency-Inverse Document Frequency (TF-IDF) matrix. The $(i, j)^{th}$ entry of this matrix represents the term frequency of token $i$ in the document $j$ normalized by the term frequency of token $i$ across all documents. Upon performing a low-rank approximation (via the Eckart-Young Theorem) to this matrix, one is able to construct low dimensional representations of these "document" column vectors thereby allowing the extraction of "latent semantics" of the corpus in a computationally efficient manner. With LSA one could, in principle, perform sentence classification, but it would clearly fail if the query text to be classified did not contain any of the dominant tokens.
\vspace{3 mm}
\\
Latent Dirichlet Allocation (LDA) [3, 4] aims to improve on this deficiency by modeling the interaction between the tokens from a probabilistic viewpoint. It works by constructing "topics" for each document: wherein each topic consists of several such dominant tokens. Words in the document can now be seen as mixture models of such topics, and the weights of these mixtures are sampled from a separate multinomial distribution. It derives its name from the fact that the parameter set for this multinomial distribution is actually derived from a Dirichlet distribution, thus generating these multivariate parameter vectors efficiently. 
\vspace{3 mm}
\\
These techniques are robust due to their independence from the requirement of working within a hypothesis framework. By utilizing the Bayesian mixture model setting in general, they come very close to theoretical guarantees of expected performance but, due to the very same architectural generality, are not suitable for fine-tuned/context-based NLU tasks. For example: LDA would be successful in retrieving closest documents matching a certain token or a sentence, but it would not tell us whether the said token is used in a positive or negative connotation.
\subsection{Recurrent Neural Networks}
While Recurrent Neural Networks themselves have been around well before the ascent of language modeling [7], the real impetus for their usage in tasks involving language translation and understanding stemmed from the construction of an Encoder-Decoder system [8]. The encoder RNN would process input tokens sequentially and concomitantly update its hidden state variable until the separator token is issued. The decoder then uses this final hidden encoder state vector to generate output tokens that maximize the conditional probability i.e., the likelihood of the source given a specific translated sentence.
\begin{figure}[!htb]
    \includegraphics[scale=0.65]{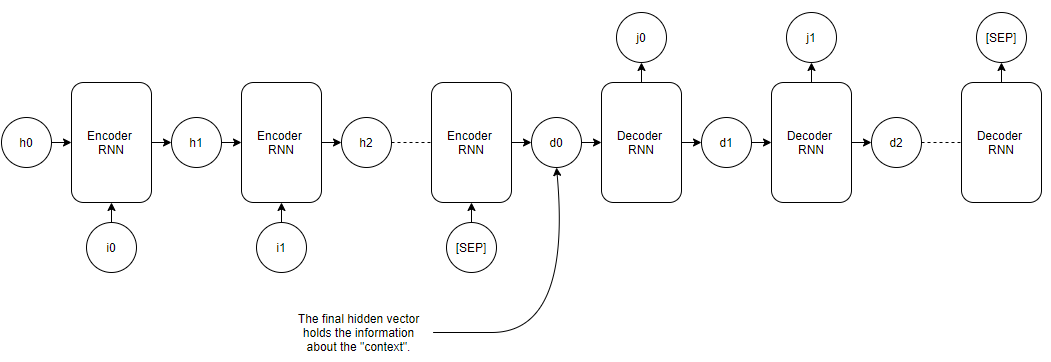}
    \caption{The Encoder-Decoder RNN Architecture}
\end{figure}
The authors of [8] subsequently propose an upgrade to the RNN cells by introducing a "gate" inside each cell that allows the $i^{th}$ cell to choose from one of the two actions: (a) update the hidden state vector coming from the $(i-1)^{st}$ cell or (b) to \textit{forget} this vector and instead reset the state back to $h_0$. This extension goes by the name of the Gated Recurrence Unit (GRU). It allows the standard RNN architecture to adapt to selective memory and makes it more agile towards handling irrelevant tokens. However, it has been definitively proven [9, 10] that GRUs significantly under-perform when compared to their more elaborate predecessor, the LSTM cell, which we now turn to. 

\subsection{Long Short-Term Memory}
The LSTM model was introduced prior to the encoder-decoder model as a multi-gate variant of the GRU. The original idea [12] started off as a "Constant-Error-Carousel" system which was designed to mitigate the impending training issues [13, 14] during backward propagation (backprop) in simple RNN/GRU based models. In short, the backprop step while training RNNs starts involving factors in the gradient that contain the weight parameters raised to the power of the memory span ($\delta t$) and depending on whether the parameter is less or greater than 1, the gradient can vanish or explode exponentially with $\delta t$. Due to the large number of such factors, this problem becomes difficult to resolve via clipping/projecting the gradients back onto a fixed domain every time such a violation takes place during automatic differentiation.
\vspace{3 mm}
\\
LSTMs solve the problem of gradient divergence by circulating the error inside the cell (hence the "carousel") using multiple stages of hidden state gating. The GRU was actually based off this initial design, wherein the authors of [8] decided to keep only the "forget" gate in the end to make the overall architecture nimble and easy to train. The template for the LSTM cell used in Figure 2 was taken from the highly recommended blog article by Chris Olah [15]. 
\vspace{3 mm}
\\
\begin{figure}[!htb]
    \includegraphics[scale=0.4]{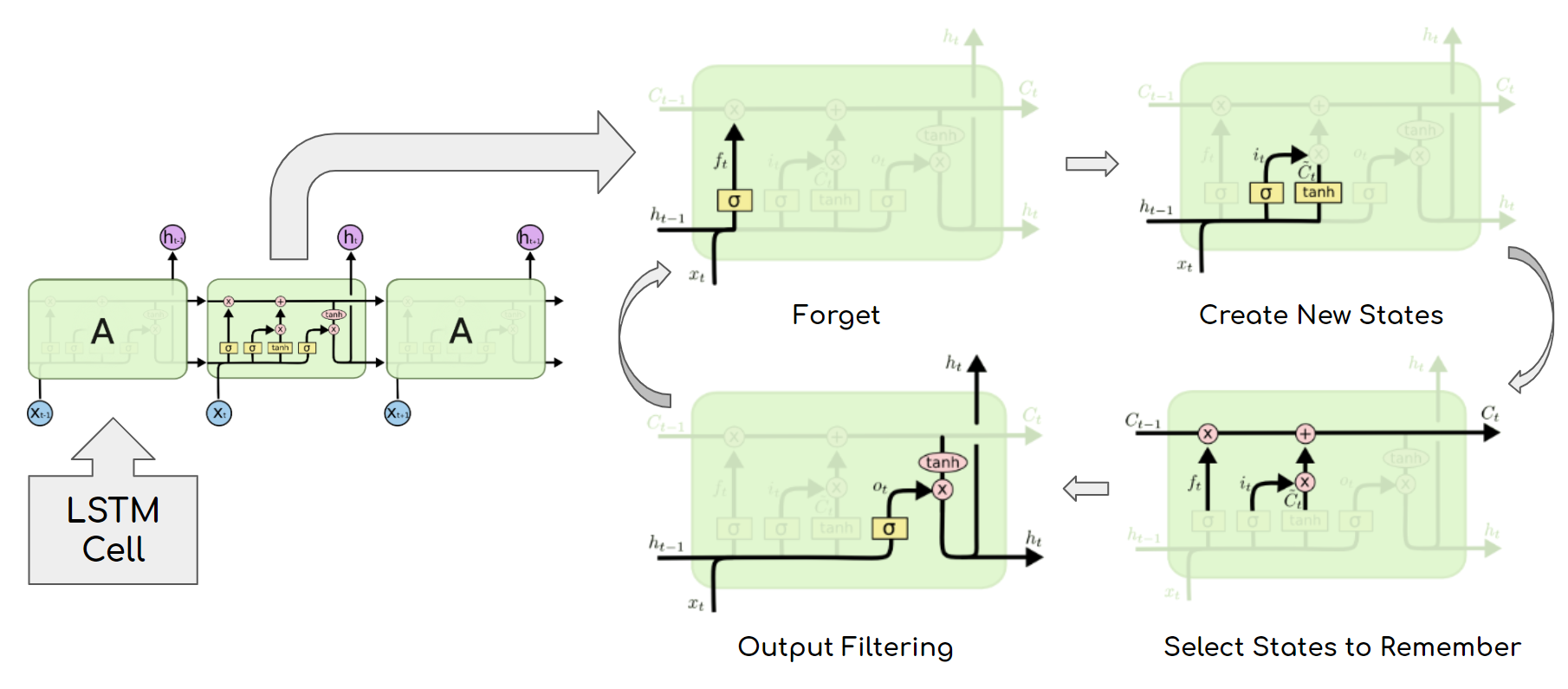}
    \caption{The LSTM State Machine}
\end{figure}
This cell structure is one of the many variants of the original model devised by Hochreiter and Schmidhuber, and [16, 17] provide another variant that allow these gates to "peep" into the processing of the other gates for more intricate dynamics. Another excellent resource especially for text generation using LSTMs is by Karpathy [18] which demonstrates the diverse applicability of these ideas: from Shakespearean playwriting to automated mathematical proof generation in \LaTeX. However, since the building blocks continue to be RNN centric, even LSTMs have the issue of divergent gradients [14]: they do not fix the problem but simply make it less likely by delaying it through error containment. Furthermore, due to the multi-gate internals which themselves comprise of smaller RNNs, training on a large document corpus by sequentially feeding in inputs is slow and impractical.

\section{Attention \& Transformers}
Notwithstanding these problems, adding further control to enable long range dependency makes LSTMs a good contender for contextual argument generation, but it faces a serious functional hazard when presented with sentences that have syntactic ambiguities. Examples of such situations, which (often unintentionally) involve multiple interpretations of a given sentence are found very frequently in literature. Take the following example that demonstrates the "Dangling Modifier" ambiguity [19]
\vspace{3 mm}
\\
\textsc{\textbf{Leafing through the pages}, the book appeared to be much more than what it initially seemed to be.}
\vspace{3 mm}
\\
For the human reader, the phrase in bold clearly refers to the narrator's action with the book as he describes its contents, but for an algorithm it may as well have been the book leafing through its own pages since it is the only noun phrase visible in the sentence. Things become complex if we have multiple noun phrases around an ambigious pronoun:
\vspace{3 mm}
\\
\textsc{He peered through the \textbf{pet door} to look at the poor \textbf{dog} and saw that \textbf{it} was visibly shivering.}
\vspace{3 mm}
\\
While the sentence may seem far too obvious from a human perspective on understanding: the dog was visibly shivering as the narrator looked inside through the pet flap, the algorithm has the propensity to correlate the pronoun "it" with the "pet door" as well, thereby wrongly concluding that the petflap was shivering instead. The takeaway here is that, not only is the causal structure of the token placement an important facet, but also establishing correlation between their relative locations becomes crucial when deriving context. It is interesting to see that this implies a significant loss of "exchangeability" when the token ordering is seen as a joint density, and thus prevents the use of the theorem due to Bruno de Finetti [20] that gave the license to probabilistic techniques such as LDA to be seen as functionally useful.
\vspace{3 mm}
\\
This idea of learning these correlation structures is known in the NMT community as "self-attention" [21] and it was effectively used to tackle the issue of long sentence parsing and translation in classic encoder-decoder RNN settings. However, the landmark paper [22] due to Vaswani et. al., shows that comparable results can be achieved by only learning these correlation structures and nothing else. That is, training based on these attention scores alone is enough for observing comparable levels of contextual relevance (thereby justifying the title of [22], which we think is cool). The block diagram of a Transformer is awfully similar to that of the LSTM and RNN based translation systems, excecpt that it now tracks all the encoder hidden states with symmetric attention span, involves positional encoding to track token locations, and uses a "masked" attention unit on the decoder side to perform translation in one go:
\begin{figure}[!htb]
    \includegraphics[scale=0.6]{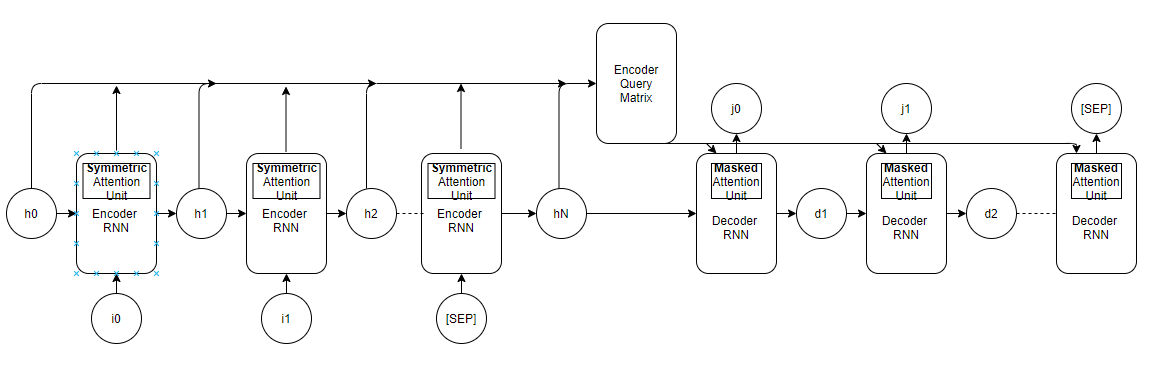}
    \caption{The Transformer}
\end{figure}
\\
The attention unit that is proposed in [22] is very intuitive in terms of how it is able to establish the dominant neighboring tokens for a given keyword. Given a token we learn a dictionary of nearby token configurations that index the corresponding likelihood probabilities/correlation coefficients. The token whose attention span is to be generated is given as a vector embedding $\bar{Q}$ to the following function:
$$\mathcal{A} = \sigma(\bar{Q}\hat{K}^T)\bar{V}$$
Here $\hat{K}$ represents the "token configuration key" matrix and $\bar{V}$ represents the vector of likelihoods. There is a stark difference between the way these attention units behave during encoding and decoding, in the sense that during encoding we can look both ways due to complete information (the "symmetric" unit: thus enabling bi-directionality) but during next token generation we mask the right half of the sentence while producing newer tokens. We have provided an infinitely rudimentary exposition of what is actually a vast subject and we refer the reader to [22, 23, 24] for finer details (such as Multi-Head Attention (MHA) and so on). 

\section{The Pre-Training Revolution: BERT \& GPT}
The development of transformer architecture has led to the evolution of many state-of-the-art language models in the natural language processing domain. BERT, which stands for Bidirectional Encoder Representation for Transformer [32], is one such revolutionary piece of technology that marks a new era in the field of Natural Language Understanding. It is made by stacking transformer encoder blocks on top of each other. BERT was an important development which managed to combine the bidirectional conditioning of each word that ELMo [31] earlier presented, along with the benefits of a fine-tunable, pre-trained transformer model. 
\vspace{3mm}
\\
The paper [32] presented two model sizes for BERT, namely BERT Base and BERT Large. The comparison between these models is presented in the following table. These models can be used for many language understanding tasks like machine translation, sentence classification, question-answering, etc. The input to any of these models is a sequence of words padded with input (token, segment and position) embeddings. Once the input is ready and is received by the  first encoder block, 'self-attention' is applied to it and the result is passed on to the feedforward network. The output of the first encoder block is then handed off to the next encoder block and the process repeats. It is important to note here that BERT does not predict the next word, rather it uses Masked Language Model (MLM) to predict random words in a particular sentence by taking into account both the right and left side context within a sentence.
\vspace{3 mm}
\\
\begin{center}
\begin{tabular}{|l|c|c|c|}
\hline
\textbf{Model }     & \multicolumn{1}{l|}{\textbf{Encoder Block Count}} & \multicolumn{1}{l|}{\textbf{Hidden Units in FF Network}} & \multicolumn{1}{l|}{\textbf{Attention Heads}} \\ \hline
BERT Base  & 12                                       & 768                                                       & 12                                   \\ \hline
BERT Large & 24                                       & 1024                                                      & 16                                   \\ \hline
\end{tabular}
\end{center}
The next innovation that followed was using only the decoder blocks of the transformer which led to the development of GPT, which stands for Generative Pre-trained Transformer [33]. GPT-1, 2 and 3 are different models based on the size of the text corpus it was trained on. Unlike BERT, these models can generate entire sentences and hence prove to be a powerful tool for language generation. GPT generates next token from a sequence of tokens in an unsupervised manner. It is auto-regressive in nature, as the newly generated token is added to the initial input sentence and the whole sentence is again provided as an input to the model. While BERT uses MLM to mask words within the entire sentence, the self-attention layer of the decoder blocks in GPT masks future words and takes into account only the past and present tokens. The processing of each token is quite similar to how it is done in BERT encoder blocks, i.e, the input includes the positional embeddings, then passes through the self-attention layer in the decoder and finally the resulting vector is passed on to the next decoder block. 

BERT and GPT are heavily pre-trained, general-purpose NLP models that bring transfer learning to the masses by allowing them to fine tune these models to their context. We have leveraged these models in our architecture to build our dialog agent in legal context.
\vspace{3 mm}
\\

\section{HumBERT}
Our architecture is inspired from the original transformer hypothesis [22] wherein we propose to expand the self-attention concept, which is currently restricted to tokens, to the larger goal of correlation in between consecutive sequences of tokens, or more generally, dialogues.
\begin{figure}[!htb]
    \includegraphics[scale=0.45]{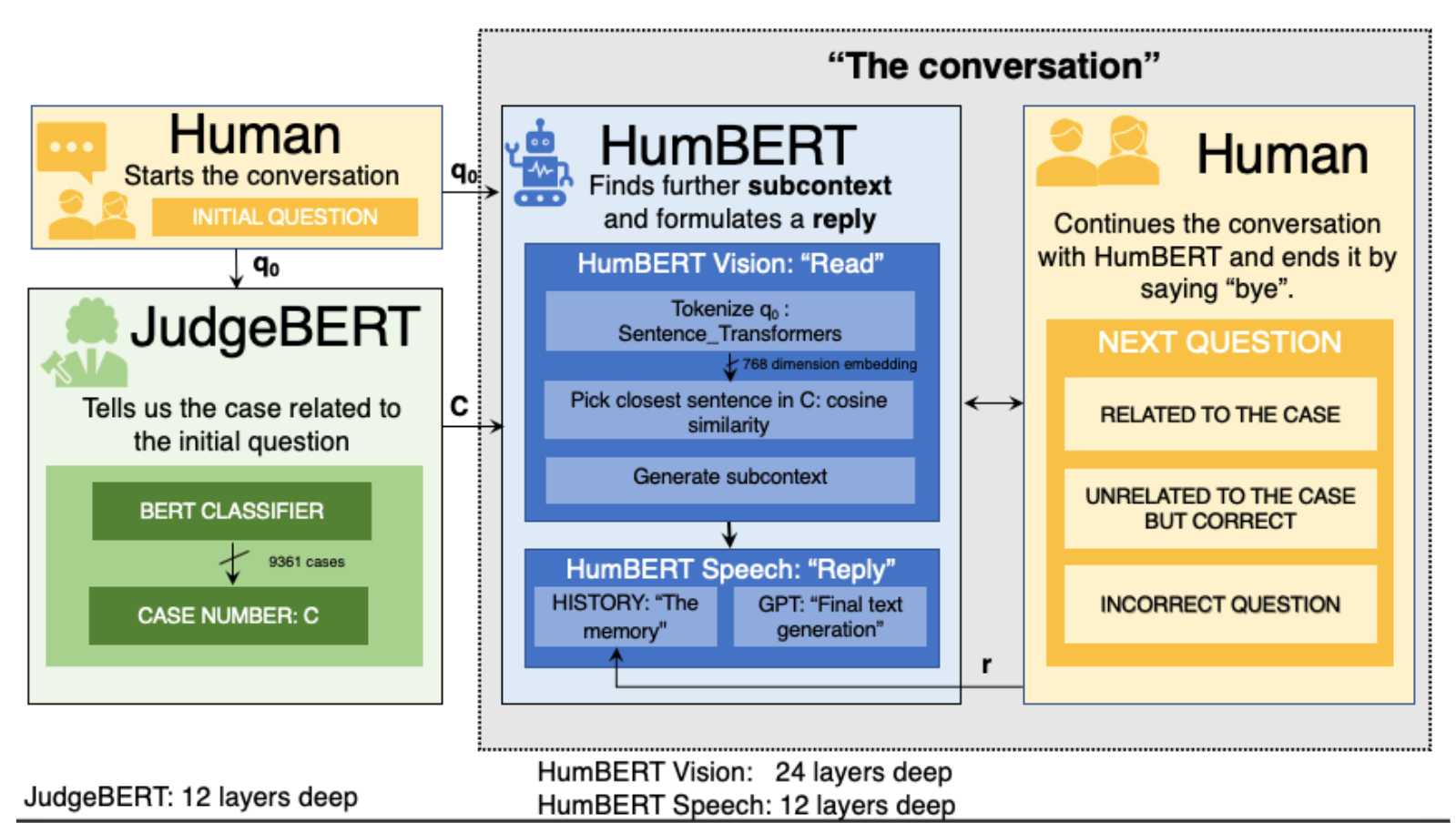}
    \caption{HumBERT Subcontext Transformer}
\end{figure}
\newpage
We believe that HumBERT can be rightfully considered as a "subcontext" driven transformer wherein it achieves three fundamental tasks:
\begin{itemize}
    \item \textbf{\underline{Seek}}: 
    \begin{itemize}
        \item Given an initial query $q_0$, the "JudgeBERT" unit $\mathcal{J}$ is activated to return which broad case file it finds $q_0$ to be most relevant to exist in.
        \item JudgeBERT is nothing but the HuggingFace BERT Base model [29] wrapped in a shallow soft-max classifier that gives us a $K$ dimensional logit vector output, where $K = ||\mu||$ and $\mu$ is the total legal corpus used for training tasks [28].
        \item We simply select the largest logit entry index and parse it as the case identifier $\mathcal{C}\in \mu$.
    \end{itemize}
      
    \item \textbf{\underline{Read}}:
    \begin{itemize}
        \item HumBERT Vision is the encoder section of the open source "Sentence Transformer" model by Reimers and Gurevych [25]. We will refer to it as $\mathcal{B}(q, \mathcal{C})$ henceforth.
        \item It is initialized on the GPU and is trained to recognize "entailment" [26] sequences in our case file. That is, if we split $\mathcal{C}$ into the sentence set $\mathcal{S} = \{S_0, S_1, ..., S_M\}$ and enforce causal ordering on $\mathcal{S}$, we consider $S_{j+1}$ to entail $S_j$ and therefore insert  $S_j \to S_{j+1}$ "entailment sequences" in our training set. Obviously, for $\mathcal{C}$ we will have $M$ such sequences.
        \item Once trained on $\mathcal{C}$, we construct the vector embedding of $\mathcal{S}$ derived by mapping the encoder on each sentence of the case $\bar{\mathcal{S}} \leftarrow \mathcal{B}(\mathcal{S}, \mathcal{C})$. We also construct the embedding for the query $\bar{q}_0 \leftarrow \mathcal{B}(q_0, \mathcal{C})$
        \item We then compute the cosine similarity [27] array $CS_j= <\bar{q}_0, \bar{S}_j>$ and find the location $j*$ which maximizes the correlation. The sentential neighborhood of $S_{j*}$ (which we call $\mathcal{N}_{j*}$) effectively represents the "subcontext" we have been alluding to thus far. 
    \end{itemize}
    
    \item \textbf{\underline{Reply}}:
    \begin{itemize}
        \item HumBERT Speech comprises of two sub-units: the conversational attention unit a.k.a the "history"/"memory" unit $\mathcal{H}$ and a response generation engine $\mathcal{G}$ which is essentially GPT-2 pretrained on $\mathcal{C}$.
        \item Once the pretraining is done, $\mathcal{G}$ processes the subcontext $\mathcal{N}_{j*}$ by using it as the seed sentence to generate $P$ contextually relevant "arguments" $\{a_1, a_2, ...., a_P\}$ . 
        \item $\mathcal{H}$ initializes itself once every conversation and holds the last $R$ "arguments" $\{h_1, h_2, ..., h_R\}$ that have been exchanged so far. Of course, when the conversation begins $\mathcal{H} = \{h_1 = q_0\}$.
        \item For each $a_l$ we calculate the average historical correlation 
        $$\rho_l = \frac{1}{R}(\sum_{k=1}^R <h_k, a_l>)$$
        \item We pick the argument that has the maximal average correlation and produce the reply $q_1$ to the original query $q_0$. We also update $\mathcal{H} \leftarrow \{q_0, q_1\}$. 
    \end{itemize}
\end{itemize}

Once $q_1$ is generated, we can cache and reuse the Vision and Speech modules $\mathcal{B}, \mathcal{G}$ to continue conversing without any additional training steps required. The following representative equations describe the iteration at a future step given human response $s_k$:
\vspace{2 mm}
\\
$$\mathcal{H} \leftarrow \mathcal{H} + \{s_k\}$$
$$j* \leftarrow arg\, max_{j = 1, 2, ..., M} <\mathcal{B}(s_k, \mathcal{C}), \bar{S}_j>$$
$$\mathcal{A} = \{a_1, a_2, ..., a_P\} \leftarrow \mathcal{G}(\mathcal{N}_{j*})$$
$$q_{k} \leftarrow max_{a_l \in \mathcal{A}}[\frac{1}{R}\sum_{h \in \mathcal{H}}<h, a_l>]$$
$$\mathcal{H} \leftarrow \mathcal{H} + \{q_k\}$$
\newpage
While our historical cache based conversational relevance scheme is primitive, it is significantly different from the approaches of both [5] and [6]: the former uses a unidirectional transformer and a policy based approach to select the apt response and the latter is suited for intent classification coupled with its matured entailment strategy. We hope to investigate better conversation patterns with different combinations of the free parameters $M, P, R$.
\section{Results}
We have trained JudgeBERT on select cases spread over 40 years of the proceedings of the Supreme Court of India [30]. We validated our model on some specimen initialization texts and our results are encouraging:
\vspace{3 mm}
\\
\textbf{\underline{Example 1}}
\vspace{1 mm}
\\
\textbf{Human: } \texttt{case involving an unjustified amount of paddy in a godown}
\vspace{2 mm}
\\
\textbf{HumBERT: } \texttt{The petitioner denied the correctness of the allegation that he had hoarded rice in an unauthorised fashion. That he took it under his thumb and took it on the spur of the moment was proved by the facts and the petitioners' testimony.}
\vspace{2 mm}
\\
\textbf{Human: } \texttt{where was this paddy and rice being smuggled to?}
\vspace{2 mm}
\\
\textbf{HumBERT: } \texttt{were smuggling rice to Meghalaya for earning undue profit? We must now find out. The answer was simple.}
\vspace{2 mm}
\\
\textbf{Human: } \texttt{what did the case simplify as a "technical error"?}
\vspace{2 mm}
\\
\textbf{HumBERT: } \texttt{In this case at least two grounds are vague, one ground is found to be false and of the remaining in one there is no explanation and in the other there is a lame excuse that the driver of the truck did not furnish the full information. He made his complaint and the court did not, in fact, accept him's evidence in all of them.}
\vspace{2 mm}
\\
\textbf{Human: } \texttt{which previous case did the attorney general and the court rely their decision upon?}
\vspace{2 mm}
\\
\textbf{HumBERT: } \texttt{v. the State of West Bengal, a Bench of five Judges of this Court explained the distinction between the vague grounds and irrelevant grounds and said that they do not stand on the same footing.}
\vspace{3 mm}
\\
\textbf{\underline{Example 2}}
\vspace{1 mm}
\\
\textbf{Human: } \texttt{cashewnut is not considered a foodstuff}
\vspace{2 mm}
\\
\textbf{HumBERT: } \texttt{When cashewnut is exported, it is exported as a foodstuff. And a foodstuff cannot, for example, be consumed by anyone who cannot produce food from them, and consequently, there could not be the same effect on the production of the food which is taken from them.}
\vspace{2 mm}
\\
\textbf{Human: } \texttt{how is a foodstuff defined}
\vspace{2 mm}
\\
\textbf{HumBERT: } \texttt{Therefore, 'foodstuff' need not necessarily mean only the final food product which is consumed. This would mean a total of 10 g/day or more. For example, a daily 2 g meal is equivalent to 2 g of sugar per day.}
\vspace{2 mm}
\\
\textbf{Human: } \texttt{how is an essential commodity defined?}
\vspace{2 mm}
\\
\textbf{HumBERT: } \texttt{What can be looked upon more of an essential commodity than both rice and paddy? It's a question that requires no further investigation. Since we can see all the other components in the rice they are much finer than either the rice or the paddy itself.}
\section{Statistics}

\begin{tabular}{|c|c|}
\hline
\textbf{Metric}                      & \textbf{Value}                                               \\ \hline
Legal Corpus Size                    & 1,383,674 Lines (39,665,578 Words)                           \\ \hline
Training Time for JudgeBERT          & 21 Hours (5.5 Hours X 4 Epochs)                              \\ \hline
Size of JudgeBERT Model              & 446 MB                                                       \\ \hline
JudgeBERT Online Resource Usage      & $\sim$1 GeForce GTX 1080 Ti GPU, 500MB, Loading Time: 9.8s   \\ \hline
JudgeBERT Case Retrieval Time        & \textless{}1s                                                \\ \hline
HumBERT Vision Online Resource Usage & $\sim$1 GeForce GTX 1080 Ti GPU, 1.2GB, \textless 5 Minutes  \\ \hline
HumBERT Speech Online Resource Usage & $\sim$2 GeForce GTX 1080 Ti GPUs, 1.2GB, \textless 5 Minutes \\ \hline
Average Query Response Time          & 32 seconds                                                   \\ \hline
\end{tabular}
\section{Future Work}
While we were successful in multi-text generation, the question remains to validate the sentence as well as paragraph level relevance. We are looking at incorporating  architectural concepts of coherence and cohesion based neural discriminators as elaborated by Cho et. al., in [34]. Simultaneously, our aim is to look into the concept of one-shot or few-shot learning to move closer towards human-like language pattern representations. 
\section*{References}
\medskip

\small

[1] Dumais, S.T., Latent semantic analysis. \textit{Annual Review of Information Science and Technology}, Vol. 38, pp. 188-230. doi:10.1002/aris.1440380105, 2004.

[2] Deerwester, S., Dumais, S. T., Furnas, G. W., Landauer, T. K., Harshman, R., Indexing By Latent Semantic Analysis, \textit{Journal of the American Society For Information Science}, 41, 391-407, 1990.

[3] Blei, D. M., Ng, Andrew., Jordan, M. I., Latent Dirichlet Allocation, \textit{Journal of Machine Learning Research}, Vol 3, pp. 993-1022, 2003.

[4] Hofmann T., Learning the Similarity of Documents : an information-geometric approach to document retrieval and categorization, \textit{Advances in Neural Information Processing Systems 12}, pp-914-920, MIT Press, 2000.

[5] Vlasov V., Mosig J. E. M., Nichol A., Dialogue Transformers, \textit{arXiv:1910.00486}, May 2020.

[6] Henderson M., Casanueva I., Mrk\v si\'c N., Su P. H., Tsung-Hsien W., Vuli\'c I., ConveRT: Efficient and Accurate Conversational Representations from Transformers,\textit{ arXiv:1911.03688}, April 2020.

[7] Rumelhart, D., Hinton, G., Williams, R., Learning representations by back-propagating errors, \textit{Nature} 323, 533–536 (1986). https://doi.org/10.1038/323533a0.

[8] Cho K., van Merrienb\"oer B., Gulcehre C., Bahdanau D., Bougares F., Schwenk H., Bengio Y., Learning Phrase Representations using RNN Encoder–Decoder for Statistical Machine Translation,\textit{ arXiv:1406.1078v3}, 2014.

[9] Weiss, G., Goldberg, Y., Yahav, E., On the Practical Computational Power of Finite Precision RNNs for Language Recognition, \textit{arXiv:1805.04908}, 2018.

[10] Britz, D., Goldie, A., Luong, Minh-Thang., Le, Q., Massive Exploration of Neural Machine Translation Architectures. \textit{arXiv:1703.03906}, 2018.

[11] Pratt L., Discriminability-Based Transfer between Neural Networks. \textit{In Advances in Neural Information Processing Systems 5, [NIPS Conference]}. Morgan Kaufmann Publishers Inc., San Francisco, CA, USA, pp. 204–211, 1992.

[12] Hochreiter S., Schmidhuber J., Long Short-Term Memory, \textit{Neural Computation}, 9(8), pp 1735-1780, 1997.

[13] Pascanu R., Mikolov T., Bengio Y., On the difficulty of training Recurrent Nueral networks, \textit{arXiv:1211.5063v2}, 2013.

[14] \textit{How does LSTM prevent the vanishing gradient problem?}, Cross Validated Stack Exchange, https://stats.stackexchange.com/q/263956.

[15] Olah C., Understanding LSTM Networks, \textit{Chris Olah's Personal Blog}, https://colah.github.io/posts/2015-08-Understanding-LSTMs/, August 2015. 

[16] Greff K., Srivastava R., Koutník J., Steunebrink B., Schmidhuber J., LSTM: A Search Space Odyssey, \textit{arXiv:1503.04069}, 2017.

[17] Graves A., Schmidhuber J., Framewise phoneme classification with bidirectional LSTM and other neural network architectures, \textit{Neural Networks}, Vol. 18(5–6), pp. 602–610, July 2005.

[18] Karpathy A., The Unreasonable Effectiveness of Recurrent Neural Networks, \textit{Andrej Karpathy Blog}, http://karpathy.github.io/2015/05/21/rnn-effectiveness/, 2015.

[19] McArthur, T., The dangling modifier or participle, \textit{The Oxford Companion to the English Language}, pp. 752-753. Oxford University Press, 1992.

[20] Diaconis, P., Freedman, D., Finite exchangeable sequences, \textit{Annals of Probability}, Vol. 8 (4), pp. 745–764, 1980.

[21] Bahdanau, D., Cho, K., and Bengio, Y., Neural Machine Translation by Jointly Learning to Align and Translate, \textit{http://arxiv.org/abs/1409.0473}, 2014.

[22] Vaswani A., Shazeer N., Parmar N., Uszkoreit J., Jones L., Gomez A., Kaiser L., Polousukhin I., Attention Is All You Need, \textit{Advances In Neural Information Processing Systems 30}, pp. 5998-6008, http://papers.nips.cc/paper/7181-attention-is-all-you-need.pdf, 2017.

[23] Kaiser L., Attentional Neural Network Models Masterclass, \textit{YouTube}, https://www.youtube.com/watch?v=rBCqOTEfxvg, October 2017.

[24] Alammar J., The Illustrated Transformer, \textit{Visualizing machine learning one concept at a time.}, http://jalammar.github.io/illustrated-transformer/, July 2018.

[25] Reimers, N., Gurevych, I., Sentence-BERT: Sentence Embeddings using Siamese BERT-Networks, \textit{Proceedings of the 2019 Conference on Empirical Methods in Natural Language Processing}, 2019.

[26] Bowman S., Angeli G., Potts C., and Manning C., A large annotated corpus for learning natural language inference, \textit{In Proceedings of the 2015 Conference on Empirical Methods in Natural Language Processing (EMNLP)}, 2015.

[27] \texttt{scipy.spatial.distance.cosine}, \textit{SciPy Documentation}, https://docs.scipy.org/doc/scipy-0.14.0/reference/generated/scipy.spatial.distance.cosine.html.

[28] McCormick C., BERT Fine-Tuning Tutorial with PyTorch, \textit{Personal Blog}, https://mccormickml.com/2019/07/22/BERT-fine-tuning/, 2019.

[29] Wolf T., Debut L., Sanh V., Chaumond J., Delangue C., Moi A., Cistac P., Rault T., Louf R., Funtowicz M., Brew J., HuggingFace's Transformers: State-of-the-art Natural Language Processing, \textit{http://arxiv.org/abs/1910.03771}, 2019.

[30] The Judgements Information System, \textit{Supreme Court of India}, www.judis.nic.in, 1950-1990.

[31] Peters, M.E., Neumann, M., Iyyer, M., Gardner, M., Clark, C., Lee, K., Zettlemoyer, L., Deep contextualized word representations. \textit{http://arxiv.org/abs/1802.05365}, 2018.

[32] Devlin J., Chang M., Lee K., Toutanova K., BERT: Pre-training of Deep Bidirectional Transformers for Language Understanding, \textit{https://arxiv.org/abs/1810.04805}, 2018.

[33] Brown, T. B., Mann, B., Ryder, N., Subbiah, M., Kaplan, J., Dhariwal, P., Agarwal, S., Language Models are Few-Shot Learners, \textit{http://arXiv.org/abs/2005.14165}, 2020.

[34] Cho, W. S., Zhang, P., Zhang, Y., Li, X., Galley, M., Brockett, C., Wang, M., Gao,J., Towards coherent and cohesive long-form text generation, \textit{https://arxiv.org/abs/1811.00511}, 2019.

\end{document}